\documentclass{article} 
\usepackage{iclr2027_conference,times}
\usepackage{graphicx}
\usepackage{float}
\usepackage{booktabs}
\usepackage{wrapfig}

\usepackage{amsmath,amsfonts,bm}

\def\eqref#1{equation~\ref{#1}}

\def\1{\bm{1}}

\DeclareMathAlphabet{\mathsfit}{\encodingdefault}{\sfdefault}{m}{sl}
\SetMathAlphabet{\mathsfit}{bold}{\encodingdefault}{\sfdefault}{bx}{n}

\usepackage{hyperref}
\usepackage{url}

\title{RespFocus: Learning Temporal Evidence for Weakly Supervised Respiratory Sound Classification}

\author{\normalfont\begin{tabular*}{\textwidth}{@{\extracolsep{\fill}}cccc@{}}
Yi Tian\textsuperscript{1,2} & Xiaohuan Chen\textsuperscript{1,2} & Yufei Tang\textsuperscript{1,2} & Xinyu Lin\textsuperscript{1,2} \\
Hongliang Li\textsuperscript{1,2} & Chengbin Hou\textsuperscript{1,*} & Degen Huang\textsuperscript{1,*} & Rongshan Yu\textsuperscript{2,*} \\
\multicolumn{4}{c}{\small \textsuperscript{1}Fuyao University of Science and Technology (fyust.edu.cn)} \\
\multicolumn{4}{c}{\small \textsuperscript{2}Xiamen University (xmu.edu.cn)} \\
\multicolumn{4}{c}{\small * Corresponding authors} \\
\multicolumn{4}{c}{\small Email: \texttt{tianyi319@stu.xmu.edu.cn}}
\end{tabular*}}

\iclrfinalcopy 
\begin{document}

\maketitle

\begin{abstract}
Respiratory auscultation provides a low-cost means of assessing pulmonary function, yet transient crackles and sustained wheezes can be difficult to recognize in noisy clinical recordings. Conventional deep learning classifiers often operate as black boxes, offering limited insight into the temporal acoustic evidence supporting their predictions. We present RespFocus, a weakly supervised framework that uses multiple instance learning (MIL) to learn inspectable temporal evidence from cycle-level labels and integrate it into classification. We also introduce PedResp2026, a multicenter pediatric respiratory sound cohort collected in routine clinical environments with adjudicated cycle-level annotations. In classification experiments, RespFocus achieves four-class Scores of $69.55\%$ on public ICBHI 2017 and $72.91\%$ on PedResp2026. Evidence diagnostics and matched temporal interventions support the functional contribution of the learned evidence to model predictions. Together, these findings highlight RespFocus's potential as a tool for automated respiratory sound diagnosis.
\end{abstract}

\begingroup
\setlength{\intextsep}{6pt}
\setlength{\abovecaptionskip}{4pt}
\setlength{\belowcaptionskip}{0pt}
\begin{figure}[H]
    \vspace{-6pt}
    \centering
    \includegraphics[width=\textwidth]{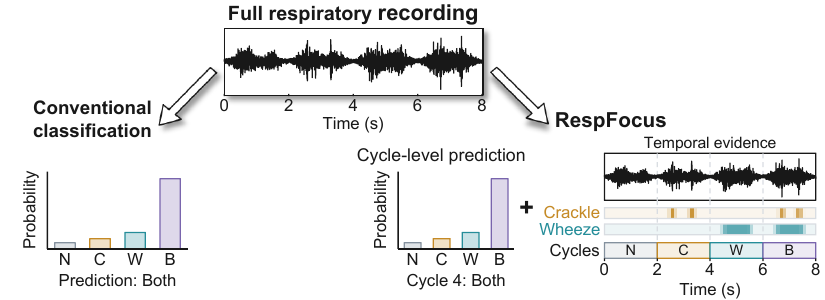}
    \caption{
    Illustrative comparison of conventional cycle classification (left) and RespFocus (right) with cycle intervals and temporal evidence. Amber/teal denote crackle/wheeze evidence. N/C/W/B denote Normal/Crackles/Wheezes/Both.
    }
    \label{fig:respfocus_concept}
    \vspace{-6pt}
\end{figure}
\endgroup

\section{Introduction}

Respiratory auscultation provides a low-cost means of assessing respiratory sounds in routine care \citep{rocha2019open,bardou2018lung,gairola2021respirenet}. Crackles typically appear as brief, discontinuous transients, whereas wheezes exhibit a more sustained tonal structure \citep{pasterkamp2016standardisation}. Automated respiratory sound classification remains challenging because these sounds differ in duration and can be obscured by speech, motion, and sensor-contact noise.

A respiratory cycle may contain crackles, wheezes, or both, and the two subtypes need not occur at the same time. Recognizing co-occurrence therefore requires gathering evidence for each subtype across the cycle. Keeping this evidence linked to the recording would also allow a listener to revisit the corresponding portions of the sound when reviewing a prediction.

Pretrained audio representations have improved respiratory sound classification \citep{bae2023patch,he2024multi,zhang2024towards}, providing a starting point for learning such evidence. However, the annotations used here specify cycle boundaries and sound categories without marking individual events. Under this weak supervision, the model must learn which temporal responses are useful by aggregating them into a prediction and comparing it with the cycle label.

Averaging across a cycle can dilute a brief crackle among less informative positions, while a single maximum captures only one response from distributed wheeze activity. A common aggregation rule therefore imposes the same temporal preference on acoustically different subtypes. This motivates separate rules that preserve crackle peaks and combine multiple wheeze responses.

We present RespFocus, a weakly supervised framework that uses multiple instance learning (MIL) \citep{dietterich1997solving,ilse2018attention} to learn inspectable temporal evidence from cycle-level labels and integrate it into classification. Each annotated cycle forms a bag of temporal instances. Building on interpretable time-series MIL \citep{early2024millet}, it combines OPERA-CT and Audio Spectrogram Transformer (AST) representations \citep{zhang2024towards,gong2021ast}. A shared abnormal-presence head guides local refinement in separate subtype branches. Crackle max pooling preserves peak responses, while wheeze top-$k$ pooling combines multiple high-scoring positions. The two subtype decisions compose the four cycle categories and represent co-occurrence within a cycle. Each prediction is accompanied by shared and subtype-specific evidence indexed to the supplied interval (Figure~\ref{fig:respfocus_concept}).

To evaluate this approach alongside ICBHI 2017, we introduce PedResp2026, a new multicenter pediatric respiratory sound dataset collected during routine care. It contains 493 recordings from 137 children at seven hospitals, with expert-adjudicated cycle labels. Patient-disjoint partitions support evaluation on held-out children from the collection. The new cohort complements the public benchmark with recordings from pediatric care and provides a resource for learning temporal evidence under weak supervision.

RespFocus improves classification on both datasets. Progressive ablations examine how performance changes as the architecture is assembled, with a controlled comparison isolating the pooling design. Matched feature-space suppression and retention test whether high-ranked evidence contributes to predictions. These experiments evaluate the classification outcome, the design choices behind it, and the role of the selected evidence. Our contributions are:
\begin{enumerate}
    \item A weakly supervised framework that uses MIL to learn inspectable temporal evidence from cycle-level labels and integrate it into respiratory sound classification.
    \item A morphology-aware aggregation and decoding design that uses crackle max pooling and wheeze top-$k$ pooling to reflect their temporal differences, then composes the two subtype decisions into four cycle categories.
    \item PedResp2026, a new multicenter pediatric respiratory sound dataset with 493 recordings from 137 children at seven hospitals, expert-adjudicated cycle labels, and patient-disjoint evaluation partitions.
    \item Evaluation on ICBHI 2017 and PedResp2026, with four-class Scores of 69.55\% and 72.91\%, respectively. Architectural ablations and matched feature-space interventions on ICBHI 2017 connect the classification gains to the proposed design and the evidence used for prediction.
\end{enumerate}

\section{Related Work}

\textbf{Weakly supervised temporal learning.}
MIL learns from bag labels without instance annotations \citep{dietterich1997solving}. Respiratory cycle annotations follow this structure, with an acoustic label for each cycle and latent locations for the supporting sounds. Attention-based MIL learns instance aggregation \citep{ilse2018attention}. MILLET extends this approach to time-series classification and local explanations \citep{early2024millet}. Both make aggregation an explicit link between local responses and a sequence-level decision. Weakly supervised action localization separates foreground activity from background \citep{paul2018wtalc,nguyen2019weakly,ren2023proposal}, while sound event detection addresses event duration and acoustic background \citep{dinkel2021duration,xin2023background}. RespFocus uses cycle labels to learn evidence within supplied intervals. Its subtype-specific pooling encodes the different temporal preferences of crackles and wheezes, and its compositional decoder preserves their co-occurrence within a bag.

\textbf{Respiratory sound representations and explanations.}
Convolutional and transformer models support respiratory sound classification \citep{bardou2018lung,gairola2021respirenet,bae2023patch,he2024multi}. Supervised audio pretraining, masked modeling, and acoustic tokenization provide transferable features \citep{kong2020panns,gong2022ssast,huang2022masked,chen2023beats}. OPERA specializes pretraining to respiratory audio \citep{zhang2024towards}. RespFocus builds on OPERA-CT and AST representations by guiding, refining, and pooling their temporal features for the two sound subtypes.

Prototype and example-based approaches make predictions interpretable through reference examples \citep{ren2022prototype,chang2022example}. RespFocus exposes temporal evidence within the input cycle and tests its contribution through interventions on the model's representations. The shared evidence describes abnormal support, and the subtype scores expose the responses pooled into the cycle-level logits. This makes the evidence accessible along the same path that produces the prediction.

\section{Methodology}
\label{sec:framework}

Given a recording $x$ and annotated cycle intervals $\{I_c\}_{c=1}^{N}$, RespFocus returns $\{(I_c,\hat y_c,S_c)\}_{c=1}^{N}$: the supplied intervals, predicted categories, and indexed temporal evidence. Each interval defines an instance bag, with predicted category $\hat y_c\in\{\mathrm{Normal},\mathrm{Crackles},\mathrm{Wheezes},\mathrm{Both}\}$. Both denotes crackles and wheezes within the same cycle, whether or not their events overlap.

\begin{figure}[t]
    \vspace{-6pt}
    \centering
    \includegraphics[width=\textwidth]{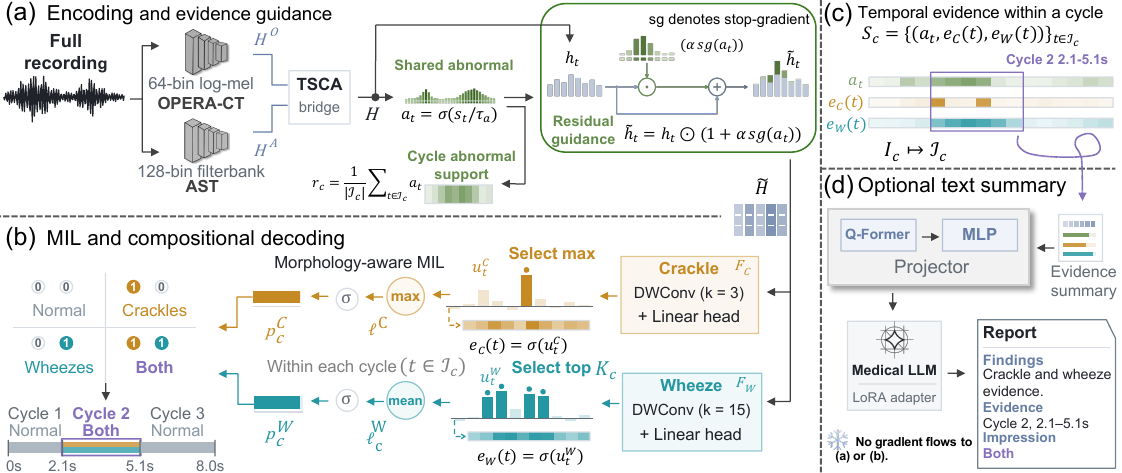}
    \caption{
    Illustrative RespFocus overview.
    (a) Dual-stream fusion and guidance from shared abnormal evidence.
    (b) Subtype MIL pooling and compositional decoding.
    (c) Shared abnormal, crackle, and wheeze evidence.
    (d) Optional text summaries with temporal references.
    }
    \label{fig:respfocus_architecture}
    \vspace{-6pt}
\end{figure}

The model fuses OPERA-aligned temporal features with AST context, uses shared abnormal evidence to guide subtype refinement, and pools the subtype logits within each cycle (Figure~\ref{fig:respfocus_architecture}). The shared head modulates temporal features according to abnormal support. The subtype branches refine these features into crackle and wheeze responses, which cycle pooling converts into the final decision. The optional text interface verbalizes the acoustic outputs.

\subsection{Dual-stream Temporal Encoding}
\label{sec:encoding}

OPERA-CT encodes a 64-bin log-mel spectrogram, while AST encodes a normalized 128-bin filterbank \citep{zhang2024towards,gong2021ast}. Writing the frontends as $\mathcal{M}_{64}$ and $\mathcal{F}_{128}$, we obtain
\begin{equation}
    \begin{aligned}
        X^O &= \mathcal{M}_{64}(x),
        & H^O &= f_O(X^O)\in\mathbb{R}^{T\times768},\\
        X^A &= \mathcal{F}_{128}(x),
        & H^A &= f_A(X^A)\in\mathbb{R}^{T_A\times768}.
    \end{aligned}
\end{equation}
Here, $T$ is the length of the OPERA-aligned temporal sequence, and $T_A$ is the number of AST tokens.

Temporal-spectral cross-attention (TSCA) projects both outputs to $d_b=1024$ and queries AST features with the OPERA-aligned sequence \citep{vaswani2017attention},
\begin{equation}
    \begin{aligned}
        Z^O &= H^O W_O,
        \qquad Z^A = H^A W_A,\\
        \hat Z^O &=
        \operatorname{LN}\!\left(
            Z^O +
            \operatorname{MHA}(Q=Z^O,K=Z^A,V=Z^A)
        \right).
    \end{aligned}
\end{equation}

A residual gate and output projection produce the shared temporal features,
\begin{equation}
    \begin{aligned}
        g^O &= \sigma\!\left(\phi_O(\hat Z^O)\right),\\
        \bar Z^O &= \hat Z^O\odot(0.1+0.9g^O),\\
        H &= \operatorname{MLP}_{\mathrm{out}}(\bar Z^O)
        = [h_1,\ldots,h_T]\in\mathbb{R}^{T\times D}.
    \end{aligned}
\end{equation}

Here, $D=512$. Each supplied interval $I_c$ maps to a nonempty token set $\mathcal I_c\subseteq\{1,\ldots,T\}$ using the nearest OPERA-token center times. Features are computed before cycle grouping. AST retains at most the first 512 input frames, so its context covers the initial segment of longer recordings. OPERA supplies the temporal indexing.

\subsection{Shared Abnormal-presence Evidence and Subtype Refinement}
\label{sec:guidance}

For each fused token $h_t$, a shared head combines local evidence and an auxiliary abnormal cue into a logit $s_t$. Its score modulates the feature through residual guidance,
\begin{equation}
    a_t=\sigma(s_t/\tau_a),\qquad
    \tilde h_t=h_t\odot\bigl(1+\alpha\,\operatorname{sg}(a_t)\bigr).
\end{equation}
Here, $\tau_a>0$ is the sigmoid temperature, $\alpha$ is the guidance strength, and $\operatorname{sg}$ stops gradients through the multiplier, broadcast across channels. The direct feature path remains trainable, and the shared head receives a separate abnormal-presence objective (Section~\ref{sec:objective}).

Inspired by convolutional local modeling \citep{gulati2020conformer,peng2022branchformer}, we refine the guided sequence $\tilde H=[\tilde h_1,\ldots,\tilde h_T]$ with separate subtype blocks, using a shorter kernel for crackles and a broader kernel for wheezes. Each block is followed by a linear subtype head,
\begin{equation}
    \begin{aligned}
        H^C &= F_C(\tilde H;k=3),
        & u_t^C &= w_C^\top h_t^C+b_C,\\
        H^W &= F_W(\tilde H;k=15),
        & u_t^W &= w_W^\top h_t^W+b_W.
    \end{aligned}
\end{equation}

Kernel sizes are in tokens and express relative temporal preferences. The subtype scores $e_C(t)=\sigma(u_t^C)$ and $e_W(t)=\sigma(u_t^W)$ form the indexed evidence collection
\begin{equation}
    S_c=\bigl\{\bigl(a_t,e_C(t),e_W(t)\bigr)\bigr\}_{t\in\mathcal I_c}.
\end{equation}
Shared evidence $a_t$ guides refinement, while $e_C(t)$ and $e_W(t)$ expose the subtype responses whose logits are pooled for classification. All three scores retain correspondence with the recording.

\subsection{Morphology-aware MIL and Compositional Decoding}
\label{sec:pooling}

Crackle pooling selects the largest logit, while wheeze pooling averages multiple high-scoring positions:
\begin{equation}
\begin{aligned}
\ell_c^C&=\max_{t\in\mathcal I_c}u_t^C,\qquad
K_c=\max\!\left(1,\left\lceil\rho_W|\mathcal I_c|\right\rceil\right),\quad\rho_W=0.35,\\
\ell_c^W&=\frac{1}{K_c}\sum_{t\in\operatorname{TopK}_{K_c}(\mathcal I_c;u^W)}u_t^W.
\end{aligned}
\end{equation}
$\operatorname{TopK}_{K_c}$ selects the $K_c$ largest wheeze logits, whose positions can be noncontiguous. The count $K_c$ differs from the convolutional kernel size $k$. Building on MIL aggregation \citep{ilse2018attention,early2024millet}, we use asymmetric pooling to preserve peak crackle responses while combining distributed wheeze evidence. Both operations produce cycle-level logits from the same temporal evidence exposed for inspection.

The pooled logits yield two subtype probabilities, which are thresholded independently using validation-selected thresholds,
\begin{equation}
    p_c^q = \sigma(\ell_c^q),
    \qquad
    \hat b_c^q = \mathbf{1}[p_c^q>\theta_q],
    \qquad q\in\{C,W\}.
\end{equation}

The ordered pair of crackle and wheeze decisions determines the category,
\begin{equation}
    \hat y_c =
    \begin{cases}
        \mathrm{Normal},
            & (\hat b_c^C,\hat b_c^W)=(0,0),\\
        \mathrm{Crackles},
            & (\hat b_c^C,\hat b_c^W)=(1,0),\\
        \mathrm{Wheezes},
            & (\hat b_c^C,\hat b_c^W)=(0,1),\\
        \mathrm{Both},
            & (\hat b_c^C,\hat b_c^W)=(1,1).
    \end{cases}
\end{equation}

The decoder assigns Both when both subtype decisions are positive. Shared abnormal evidence guides feature refinement upstream of this decision.

\subsection{Training Objective}
\label{sec:objective}

Let $y_c^C,y_c^W\in\{0,1\}$ encode the presence of each subtype, and let $y_c^A=\mathbf 1[y_c^C\lor y_c^W]$ denote abnormal presence. The primary objective applies asymmetric loss (ASL) to the MIL-pooled subtype logits \citep{ridnik2021asymmetric},

\begin{equation}
\mathcal L_{\mathrm{pair}}=\frac{1}{N}\sum_{c=1}^N\operatorname{ASL}\!\left((\ell_c^C,\ell_c^W),(y_c^C,y_c^W)\right).
\end{equation}

The auxiliary objective trains the shared head to recognize abnormal support across both subtypes. It averages the shared evidence logits within each cycle and applies a binary abnormal-presence loss,

\begin{equation}
\begin{aligned} \bar s_c&=\frac{1}{\lvert\mathcal I_c\rvert}\sum_{t\in\mathcal I_c}s_t,\\ \mathcal L_{\mathrm{abn}}&=\frac{1}{N}\sum_{c=1}^N\operatorname{BCEWithLogits}(\bar s_c,y_c^A).\end{aligned}
\end{equation}

The acoustic model is optimized with

\begin{equation}
\mathcal L=\mathcal L_{\mathrm{pair}}+\omega_{\mathrm{abn}}\mathcal L_{\mathrm{abn}},\qquad\omega_{\mathrm{abn}}=1.
\end{equation}

Both labels activate both subtype targets, while Normal labels set both to zero. Each subtype branch can therefore learn from cycles containing its subtype alone or together with the other subtype. The losses supervise temporal responses through cycle aggregation, allowing evidence learning without event-level annotations.

For evidence analysis and presentation, we separately summarize the shared scores as

\begin{equation}
r_c=\frac{1}{\lvert\mathcal I_c\rvert}\sum_{t\in\mathcal I_c}a_t.
\end{equation}

Thus, analysis uses the mean probability $r_c$, whereas training uses the mean logit $\bar s_c$.

\section{Experimental Setup}

\subsection{Datasets}
We evaluate respiratory sound classification on ICBHI 2017 \citep{rocha2019open} and PedResp2026. Our fixed ICBHI partition uses 777/90/53 recordings and 5,815/657/426 cycles for training, validation, and testing. Test recordings follow the OPERA held-out list.

PedResp2026 is a newly collected pediatric cohort comprising 493 recordings from 137 children at seven hospitals, with nine device identifiers. Children aged 0.1--13 years were recorded in routine clinical rooms using an eight-site thoracic protocol. Two physicians annotated cycles, with adjudication by an additional physician from an eight-physician panel. The patient-disjoint training, validation, and test sets contain 109/14/14 children, 394/49/50 recordings, and 899/119/117 respiratory cycles, respectively.
\subsection{Baselines and Implementation}
We compare with respiratory sound classifiers using convolutional, transformer, contrastive, masked-audio, and adaptation-based approaches \citep{moummad2023pretraining,bae2023patch,kim2023adversarial,kim2024bts,kim2024stethoscope,niizumi2024masked,he2024multi,dong2025low,dong2025respiratory}.
We also consider general MIL baselines for time-series classification \citep{early2024millet,pmlr-v235-chen24af}, anomaly detection \citep{10.1609/aaai.v39i17.33933}, and instance aggregation \citep{Tang_2024_CVPR}, alongside an interpretable mixture-of-experts model \citep{Xie2026AnchorMoEIT}, for adaptation to respiratory sound classification.

OPERA-CT uses 16-kHz audio, 64 mel bins, a 2,048-sample FFT/window, and a 512-sample hop. AST uses a normalized 128-bin Kaldi filterbank, a 10-ms frame shift, and 512 input frames. We train with AdamW and cosine learning-rate decay over five independent seeds. Checkpoints and subtype thresholds are selected on validation data and fixed for testing. The wheeze pooling ratio is $\rho_W=0.35$, and the crackle and wheeze kernels are 3 and 15 tokens.
\subsection{Evaluation Metrics}

\textbf{Classification.} We use the ICBHI challenge metric definitions \citep{rocha2019open} on the study partitions above. Let $M_{ij}$ count cycles with reference class $i$ and predicted class $j$, where $N,C,W,B$ denote Normal, Crackles, Wheezes, and Both. Specificity (Sp) measures correct normal classification. Sensitivity (Se) measures correct abnormal-subtype classification, penalizing both missed abnormalities and subtype confusions:
\begin{equation}
\begin{aligned}
\mathrm{Sp}
&= \frac{M_{NN}}{\sum_j M_{Nj}}, \\
\mathrm{Se}
&= \frac{M_{CC}+M_{WW}+M_{BB}}
{\sum_{i\in\{C,W,B\}}\sum_j M_{ij}}, \\
\mathrm{Score}
&= \tfrac{1}{2}(\mathrm{Sp}+\mathrm{Se}).
\end{aligned}
\end{equation}
We report Sp, Se, and Score as percentages, alongside macro-AUROC and macro-F1.

\textbf{Temporal evidence.} We evaluate two properties of the shared evidence. Abnormal-presence AUROC uses $r_c$ to rank cycles labeled Normal versus Abnormal. CycleHitRate measures the fraction of abnormal cycles containing at least one token above a validation-selected evidence threshold,
\begin{equation}
\operatorname{CycleHitRate}=\frac{1}{|\mathcal C_{\mathrm{abn}}|}\sum_{c\in\mathcal C_{\mathrm{abn}}}\mathbf 1\!\left[\max_{t\in\mathcal I_c}a_t\geq\tau_{\mathrm{hit}}\right].
\end{equation}
Here, $\mathcal C_{\mathrm{abn}}$ contains abnormal cycles. We test functional relevance by suppressing or retaining selected tokens under matched feature-space budgets, including random, low-evidence, and energy-matched controls.

\section{Results}
\subsection{Classification performance}
RespFocus achieves the highest mean four-class Score among the compared methods on both datasets (Table~\ref{tab:resp_classification}). Scores of 69.55\% on ICBHI and 72.91\% on PedResp2026 exceed the strongest baselines, MVST (66.56\%) and Patch-Mix CL (68.41\%), by 2.99 and 4.50 percentage points. RespFocus also achieves the highest sensitivity, improving by 4.92 and 16.06 points over these baselines. On PedResp2026, sensitivity rises from 52.12\% to 68.18\%, showing stronger abnormal-subtype recognition.
\begin{table}[H]
\vspace{-6pt}
\centering
\caption{Four-class classification (\%). Sc denotes Score. RespFocus reports five-run means. Bold/underlined values mark best/second-best results.}
\label{tab:resp_classification}
\small
\setlength{\tabcolsep}{2.5pt}
\renewcommand{\arraystretch}{0.95}
\resizebox{\linewidth}{!}{%
\begin{tabular}{@{}lllcccccc@{}}
\toprule
Method & Backbone & Venue & \multicolumn{3}{c}{ICBHI 2017} & \multicolumn{3}{c}{PedResp2026} \\
\cmidrule(lr){4-6}
\cmidrule(lr){7-9}
 & & & Sc & Se & Sp & Sc & Se & Sp \\
\midrule
CE \citep{moummad2023pretraining} & CNN6 & WASPAA'23 & 55.19 & 40.32 & 70.07 & 50.00 & 20.00 & 80.00 \\
SCL \citep{moummad2023pretraining} & CNN6 & WASPAA'23 & 57.57 & 39.21 & 75.93 & 61.92 & 46.97 & 76.86 \\
AFT on Mixed \citep{kim2023adversarial} & AST & NeurIPS W'23 & 61.80 & 42.86 & 80.73 & 65.31 & 47.88 & 82.75 \\
Fine-tuning \citep{bae2023patch} & AST & INTERSPEECH'23 & 59.52 & 41.90 & 77.13 & 57.08 & 27.88 & \underline{86.27} \\
Patch-Mix CL \citep{bae2023patch} & AST & INTERSPEECH'23 & 62.34 & 43.02 & 81.67 & \underline{68.41} & \underline{52.12} & 84.71 \\
Audio-CLAP \citep{kim2024bts} & CLAP & INTERSPEECH'24 & 62.73 & 44.60 & 80.87 & 65.24 & 49.70 & 80.78 \\
BTS \citep{kim2024bts} & CLAP & INTERSPEECH'24 & 63.56 & 45.71 & 81.40 & 64.83 & 48.48 & 81.18 \\
DAT \citep{kim2024stethoscope} & AST & ICASSP'24 & 59.84 & 42.54 & 77.13 & 64.13 & 43.94 & 84.31 \\
SG-SCL \citep{kim2024stethoscope} & AST & ICASSP'24 & 61.68 & 43.49 & 79.87 & 64.73 & 51.82 & 77.65 \\
M2D-X \citep{niizumi2024masked} & M2D ViT & TASLP'24 & 63.31 & 45.08 & 81.53 & 62.50 & 46.97 & 78.04 \\
MVST \citep{he2024multi} & AST & ICASSP'24 & \underline{66.56} & \underline{51.11} & 82.00 & 54.44 & 28.48 & 80.39 \\
Low-rank \citep{dong2025low} & LoRA & ICASSP'25 & 60.68 & 36.03 & \textbf{85.33} & 56.76 & 16.67 & \textbf{96.86} \\
Single-view \citep{dong2025respiratory} & AST & BSPC'25 & 66.03 & 49.52 & 82.53 & 51.75 & 38.79 & 64.71 \\
\midrule
RespFocus & Dual-stream & This work & \textbf{69.55} & \textbf{56.03} & \underline{83.07} & \textbf{72.91} & \textbf{68.18} & 77.65 \\
\bottomrule
\end{tabular}%
}
\vspace{-6pt}
\end{table}

\begin{figure}[t]
\vspace{-6pt}
\centering
\includegraphics[width=\linewidth]{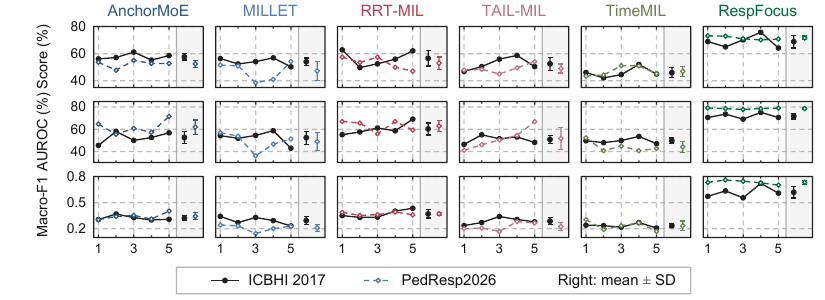}
\caption{Five-run results by metric (rows) and method (columns) for ICBHI 2017 (solid black lines, filled circles) and PedResp2026 (colored dashed lines, open diamonds). Right summaries show mean $\pm$ sample SD. Run indices do not imply paired initialization across methods.}
\label{fig:temporal_baselines}
\vspace{-6pt}
\end{figure}

RespFocus also outperforms five adapted MIL and temporal baselines in mean Score, macro-AUROC, and macro-F1 on both datasets. The strongest baselines by Score are AnchorMoE \citep{Xie2026AnchorMoEIT} on ICBHI and RRT-MIL \citep{Tang_2024_CVPR} on PedResp2026. RespFocus exceeds their Scores by 11.91 and 19.85 percentage points, respectively. Compared with RRT-MIL, the strongest macro-F1 baseline on both datasets, RespFocus raises macro-F1 from 0.371 to 0.625 on ICBHI and from 0.372 to 0.744 on PedResp2026. These gains show that the advantage extends to performance averaged equally across the four classes. Figure~\ref{fig:temporal_baselines} shows the variation across five runs, with RespFocus Score SDs of 4.92 points on ICBHI and 2.42 on PedResp2026. We next examine the design choices within RespFocus through ablations, complementing these comparisons of complete classification pipelines.

\subsection{Architectural ablations}
Adding temporal encoding, evidence guidance, and subtype decisions progressively improves ICBHI classification (Table~\ref{tab:progressive_ablation}). The mean ICBHI Score rises from 59.79\% for a flat AST classifier to 62.40\% with dual-stream TSCA, 64.66\% with the evidence pathway, and 66.81\% with binary-pair decoding and mean pooling. The last comparison holds the remaining architecture and training protocol fixed: replacing mean pooling in both subtype branches with crackle max pooling and wheeze top-$k$ pooling raises Score to 69.55\%, a 2.74-point gain. The pooling change raises sensitivity from 50.95\% to 56.03\%, while specificity changes from 82.67\% to 83.07\%. Most of the Score increase therefore comes from correct abnormal-subtype predictions. Since this sensitivity metric rewards the correct Crackles, Wheezes, or Both category, the result links the aggregation change to recognition of the sounds present within abnormal cycles.
\begin{table}[t]
\vspace{-6pt}
\centering
\caption{Progressive ICBHI 2017 four-class ablation (\%, five-run mean $\pm$ SD).}
\label{tab:progressive_ablation}
\small
\resizebox{\linewidth}{!}{%
\begin{tabular}{lccccccc}
\toprule
Variant & TSCA & Evidence & Binary pair & Asym. pooling & Sp (\%) & Se (\%) & Score (\%) \\
\midrule
Flat AST classifier & $\times$ & $\times$ & $\times$ & $\times$ & $76.40\pm6.35$ & $43.17\pm4.07$ & $59.79\pm4.47$ \\
Dual-stream global pooling & $\checkmark$ & $\times$ & $\times$ & $\times$ & $79.40\pm5.67$ & $45.40\pm3.48$ & $62.40\pm4.11$ \\
Evidence pathway + flat four-way & $\checkmark$ & $\checkmark$ & $\times$ & $\times$ & $81.07\pm5.91$ & $48.25\pm3.52$ & $64.66\pm4.28$ \\
Mean pooling & $\checkmark$ & $\checkmark$ & $\checkmark$ & $\times$ & $82.67\pm5.48$ & $50.95\pm3.14$ & $66.81\pm2.85$ \\
Full RespFocus & $\checkmark$ & $\checkmark$ & $\checkmark$ & $\checkmark$ & $83.07\pm8.50$ & $56.03\pm4.07$ & $69.55\pm4.92$ \\
\bottomrule
\end{tabular}%
}
\end{table}

\subsection{Functional role of temporal evidence}
\label{sec:temporal_evidence}
The shared evidence distinguishes abnormal cycles and contributes to feature refinement. Its cycle summary $r_c$ achieves an AUROC of $(83.51\pm0.48)\%$ for separating abnormal from normal cycles. To test its role in refinement, we perturb the shared guidance scores $a_t$ at a fixed checkpoint. Replacing the scores with the recording mean removes temporal variation and reduces Score from 70.02\% to 65.78\%. Removing guidance lowers Score to 62.63\%. Within-cycle shuffling preserves each cycle's score distribution but changes its temporal assignment, yielding a Score of 68.28\%. These decreases show that the classifier uses the shared guidance, including its temporal arrangement.

Matched feature-space interventions then assess the contribution of the selected representations. Under equal token budgets, suppressing high-evidence tokens reduces mean ICBHI Score by 11.13 percentage points, compared with 4.81 for random tokens and 6.23 for energy-matched random tokens. Retaining only high-evidence tokens yields a Score of 65.12\%, versus 55.17\% for random retention. The larger suppression effect and stronger retention performance jointly support the evidence ranking as a way to identify features that contribute to classification.

\begin{figure}[t]
\centering
\includegraphics[width=\linewidth]{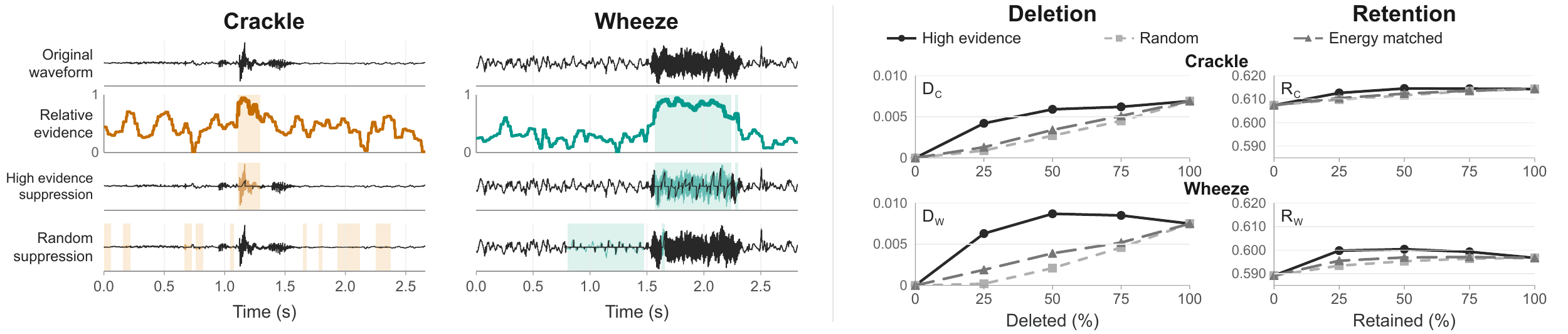}
\caption{Illustrative waveform interventions with within-cycle min--max-normalized evidence (left).
Right panels show deletion $D_q(b)=p_q(x)-p_q(x_{\mathrm{delete},b})$ and retention $R_q(b)=p_q(x_{\mathrm{retain},b})$ for subtype $q\in\{C,W\}$ and cycle-duration fraction $b$.}
\label{fig:evidence_intervention}
\vspace{-18pt}
\end{figure}

Figure~\ref{fig:evidence_intervention} illustrates complementary interventions on waveform windows ranked by subtype evidence. On the left, normalized crackle evidence peaks near a brief transient, while wheeze evidence remains elevated over a longer interval. The highlighted windows show where high-evidence and random suppression modify the waveforms. On the right, deleting high-evidence windows causes larger drops in subtype probability than random or energy-matched deletion at partial budgets. Retaining high-evidence windows also yields higher subtype probabilities at small and intermediate budgets. The curves converge at the full-cycle budget, when all strategies delete or retain the same samples. These examples connect the temporal evidence profiles to the prediction changes caused by selecting different portions of the waveform.

\begin{wrapfigure}{R}{0.5\textwidth}
    \centering
    \includegraphics[width=\linewidth]{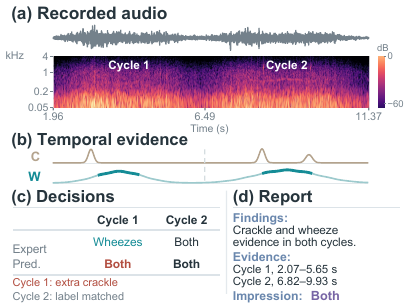}
    \caption{Two-cycle illustration of a wheeze-dominant case.}
    \label{fig:wheeze_case_compact}
\end{wrapfigure}
Figure~\ref{fig:wheeze_case_compact} illustrates two cycles from a wheeze-dominant case. RespFocus correctly predicts Cycle~2 as Both but also assigns Both to the Wheezes-only Cycle~1. The extra crackle prediction propagates into the illustrative text panel.

\section{Discussion}

RespFocus connects weakly supervised respiratory sound classification with inspection of the temporal evidence used for each decision. The classification results (Table~\ref{tab:resp_classification}) and pooling ablation (Table~\ref{tab:progressive_ablation}) support using sound morphology to guide how evidence is accumulated. This allows brief crackle responses and longer wheeze activity to contribute through different aggregation rules, while compositional decoding represents their co-occurrence within a cycle. The matched interventions show that the evidence ranking identifies representations on which predictions depend. Because these representations retain their temporal indices, the predicted category can be examined alongside the responses that produced it. This gives a listener a concrete starting point for reviewing a prediction and investigating an incorrectly assigned subtype. PedResp2026 provides a new multicenter pediatric setting for studying this connection, with adjudicated recordings from routine care and evaluation on held-out children. The optional text interface conditions generation on acoustic predictions, subtype probabilities, and cycle intervals. This conditioning provides an explicit reference for generation and can reduce the risk of language model hallucinations. The extra crackle statement in Figure~\ref{fig:wheeze_case_compact} reflects an error inherited from the acoustic model.

\paragraph{Limitations.}
PedResp2026 remains limited in size, and future work will expand the cohort to include more patients and a wider range of respiratory sounds. Expert event annotation is also labor-intensive, particularly for brief crackles that require millisecond-scale temporal precision. Learning from cycle-level labels reduces this annotation burden, although evaluating event localization still requires fine-grained reference annotations. The current patient-disjoint split leaves generalization to unseen hospitals and devices open. Future evaluations should test these acquisition-domain shifts \citep{koh2021wilds}. Evaluation of the text interface should distinguish errors propagated from the acoustic model from unsupported content introduced during text generation. It should also assess how conditioning affects hallucination risk and whether the generated text helps listeners review recordings.

\section{Conclusion}
We presented RespFocus, a weakly supervised framework that uses MIL to learn temporal evidence from cycle-level labels and integrate it into respiratory sound classification. Shared abnormal evidence guides separate crackle and wheeze branches, whose pooling rules preserve brief peaks and combine distributed responses before compositional decoding determines the cycle category. Evaluation on ICBHI 2017 and PedResp2026, our new multicenter pediatric cohort, shows improved classification, while matched interventions support the contribution of the learned evidence to predictions. These findings support learning temporal evidence as part of the classifier, connecting each predicted sound category to the temporal responses used to produce it.

\subsection*{AI use statement}

We used GPT-6 Astra for language polishing, literature search, manuscript-structure review, and checks of numerical consistency and claim--evidence alignment. The authors take full responsibility for the manuscript's content, including its text, references, and scientific claims.

\subsection*{Ethics statement}

The PedResp2026 study received institutional ethics approval. Written informed consent was obtained from legal guardians, with child assent where appropriate. The data were de-identified before analysis to protect participant privacy. We plan to release PedResp2026 through controlled access upon acceptance to support reproducibility and further research.

\subsection*{Reproducibility statement}

Our code and dataset access information are available at \url{https://anonymous.4open.science/r/RespFous-31C6/}.



\bibliography{iclr2027_conference}
\bibliographystyle{iclr2027_conference}

\end{document}